\documentclass{article}

     \usepackage[nonatbib,preprint]{neurips_2019}

\usepackage[utf8]{inputenc} % allow utf-8 input
\usepackage[T1]{fontenc}    % use 8-bit T1 fonts
\usepackage{hyperref}       % hyperlinks
\usepackage{url}            % simple URL typesetting
\usepackage{booktabs}       % professional-quality tables
\usepackage{amsfonts}       % blackboard math symbols
\usepackage{nicefrac}       % compact symbols for 1/2, etc.
\usepackage{microtype}      % microtypography
\usepackage{amsmath}	% required for `\bmatrix' (yatex added)
\usepackage{amsthm}
\usepackage{algorithm}
\usepackage{algorithmic}
\usepackage{graphicx}
\usepackage{subfig}

\def\etal{\rm et al. }
\def \figref  #1{Figure \ref{#1}}
\def \eqref  #1{Eq. (\ref{#1})}
\def \tabref  #1{Table \ref{#1}}
\def \secref  #1{Section \ref{#1}}
\def \algref #1{Algorithm \ref{#1}}

\newtheorem{lemm}{Lemma}%[section]
\newtheorem{prop}{Proposition}%[section]

\title{Compact Approximation for Polynomial of Covariance Feature}

\author{%
  Yusuke Mukuta \\
  The University of Tokyo, RIKEN AIP\\
  \texttt{mukuta@mi.t.u-tokyo.ac.jp} \\
  \And
  Tatsuaki Machida \\
  The University of Tokyo\\
  \texttt{machida@mi.t.u-tokyo.ac.jp} \\
  \And
  Tatsuya Harada \\
  The University of Tokyo, RIKEN AIP\\
  \texttt{harada@mi.t.u-tokyo.ac.jp} \\
}

\begin{document}

\maketitle

\begin{abstract}
Covariance pooling is a feature pooling method with good classification accuracy.
Because covariance features consist of second-order statistics, the scale of the feature elements are varied.
Therefore, normalizing covariance features using a matrix square root affects the performance improvement.
When pooling methods are applied to local features extracted from CNN models, the accuracy increases when the pooling function is back-propagatable and the feature-extraction model is learned in an end-to-end manner.
Recently, the iterative polynomial approximation method for the matrix square root of a covariance feature was proposed, and resulted in a faster and more stable training than the methods based on singular-value decomposition.
In this paper, we propose an extension of compact bilinear pooling, which is a compact approximation of the standard covariance feature, to the polynomials of the covariance feature. 
Subsequently, we apply the proposed approximation to the polynomial corresponding to the matrix square root to obtain a compact approximation for the square root of the covariance feature.
Our method approximates a higher-dimensional polynomial of a covariance by the weighted sum of the approximate features corresponding to a pair of local features based on the similarity of the local features.
We apply our method for standard fine-grained image recognition datasets and demonstrate that the proposed method shows comparable accuracy with fewer dimensions than the original feature.
\end{abstract}

\section{Introduction}
Feature pooling is a method to summarize the statistics of local features extracted from one image into one global feature.
Earlier, feature pooling was applied to handcrafted features such as SIFT \cite{lowe2004distinctive} and HOG \cite{dalal2005histograms}.
Currently, feature pooling methods are applied to activate the convolutional layers of convolutional neural networks (CNNs) such as VGG-Net \cite{Simonyan15} and ResNet \cite{he2016deep} to improve classification accuracy.

Among the feature pooling methods, bilinear pooling \cite{lin2015bilinear} that uses second-order statistics as a global feature demonstrates good accuracy, and various extensions of bilinear pooling have been proposed.

One problem of the original bilinear pooling is its dimension. The dimensionality of bilinear pooling is the squared order of the dimensionality of local features. This is problematic particularly when bilinear pooling is applied to CNN features because they tend to be high-dimensional. To reduce the feature dimension, compact bilinear pooling \cite{gao2016compact} exploits the tensor sketch \cite{pham2013fast} and random Macraulin \cite{kar2012random} approximations for polynomial kernels to construct a differentiable pooling layer.
Compact bilinear pooling demonstrates a comparable accuracy to bilinear pooling with fewer dimensions.

Another problem is that the feature is affected by the elements of local features with large values because bilinear pooling uses second-order information. Improved bilinear pooling \cite{lin2017improved} applies matrix square root normalization to alleviate this problem. Matrix square root normalization is a method that uses 1/2 power of a bilinear feature as a global feature. Because a bilinear feature is positive semi-definite, we can calculate the matrix square root normalization by first applying singular-value decomposition (SVD) and subsequently multiplying the singular matrices with 1/2 power of the singular values. The original improved bilinear pooling and its variant \cite{li2017second} calculate the backpropagation of matrix square root normalization using matrix backpropagation \cite{ionescu2015matrix} through the differential with respect to the singular values and singular vectors. These methods exhibit slowness and instability in SVD calculations on a GPU. Furthermore, when  different singular values have the same value, the decomposition is not unique and, thus, we cannot compute the differential, thus resulting in the failure of computation of backpropagation.
Recently, the iSQRT-COV \cite{li2018towards} proposed avoiding SVD calculation using the approximation of 1/2 power of matrices with Newton--Schulz Iteration \cite{higham2008functions}. Newton--Schulz Iteration approximates the 1/2 power of matrices with only the summation and multiplication of matrices; thus, this method is compatible with the GPU computation and CNN learning framework. Furthermore, the approximated matrices are the polynomials of the input matrices. iSQRT-COV demonstrated good recognition accuracy with fast computation.

\begin{figure}
\centering
\includegraphics[width=\hsize]{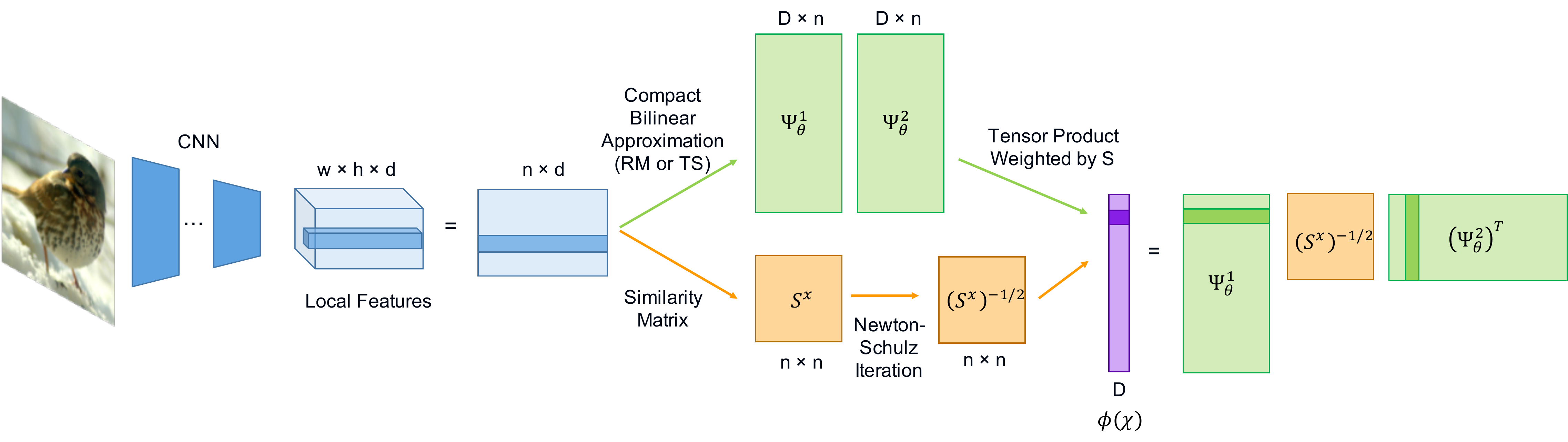}
\caption{Overview of the proposed feature coding method. We first extract CNN local feature $\mathcal{X}=\{x_n\}_{n=1}^{n_x}$. We subsequently calculate $\Psi^i\left(\mathcal{X}\right)$, $S^x$ and $\left(S^x\right)^{-1/2}$. We finally calculate $\left(S^x\right)^{-1/2}$-weighted sum of products of $\Psi^1\left(\mathcal{X}\right)$ and $\Psi^2\left(\mathcal{X}\right)$ as a global feature.}
\label{fig:system}
\end{figure}

Based on these studies, we propose a novel extension of the bilinear pooling layer that is compact, normalized, and easy to calculate backpropagation. Hence, we propose a compact approximation for the bilinear feature with iterative matrix square root normalization.
Because matrix square root normalization can be approximated using a polynomial, we first propose an approximation for a general polynomial of the bilinear feature. 
We calculate the approximation by first representing the polynomial of a covariance using the weighted sum of a bilinear feature that corresponds to a pair of local features, and subsequently calculate the weight and approximation for each bilinear feature.
We then apply the proposed method to the polynomial corresponding to the 1/2 power to obtain the approximate feature for matrix square root normalization. Because we can use Newton--Schulz Iteration to calculate the weights of summation, we avoid SVD for our approximation method. We plot the overview of the proposed method in \figref{fig:system}.

We applied the proposed approximation method to ResNet and evaluated the methods on standard fine-grained image recognition datasets.
Our method exhibited better accuracy than the existing compact approximation methods and demonstrated comparable accuracy with the original normalized covariance features.

Our contributions are as follows:
\begin{itemize}
\item We proposed a novel extension of bilinear pooling that approximates the square root of the covariance feature with low dimension via Newton--Schulz Iteration.
\item We evaluated the proposed method on standard fine-grained image recognition datasets and confirmed that our method demonstrates comparable performance to the iSQRT-COV with lower dimension and with lower computation time than Monet.
\end{itemize}

\begin{algorithm}[t]
\caption{Calculate tensor sketch approximation \cite{gao2016compact}}         
\label{alg:ts}                          
\begin{algorithmic}                  
\REQUIRE $\mathcal{X}=\{x_n\}_{n=1}^{n_x}$, where $x_n\in\mathbb{R}^d$, feature dimension $D$
\ENSURE $\phi^{\mathrm{TS}}(\mathcal{X}) \in \mathbb{R}^D$
\STATE Generate uniform random $h_k \in \{1, 2, ..., D\}^d$ and $s_k \in \{1, -1\}^d$ for $k=1, 2$
\STATE Define $\psi^{\mathrm{TS}}(x,h,s) = \{\left(Q x\right)_1, ..., \left(Q x\right)_D\}$ where $\left(Qx\right)_j = \sum_{\{i\in \{1, 2, ..., d\} | h(i)=j\}}{s(i)x^i}$
\STATE $\phi^{\mathrm{TS}}(\mathcal{X})=\sum_{n=1}^{n_x} \mathrm{fft}^{-1}\left(\mathrm{fft}\left(\psi^{\mathrm{TS}}(x_n,h_1,s_1)\circ\left(\psi^{\mathrm{TS}}(x_n,h_2,s_2)\right)\right)\right)$
\end{algorithmic}
\end{algorithm}

\begin{algorithm}[t]
\caption{Calculate random Maclaurin approximation \cite{gao2016compact}}         
\label{alg:rm}                          
\begin{algorithmic}                  
\REQUIRE $\mathcal{X}=\{x_n\}_{n=1}^{n_x}$, where $x_n\in\mathbb{R}^d$, feature dimension $D$
\ENSURE $\phi^{\mathrm{RM}}(\mathcal{X}) \in \mathbb{R}^D$
\STATE Generate uniform random $W_k \in \{1, -1\}^{D\times d}$ for $k=1, 2$
\STATE $\phi^{\mathrm{RM}}(\mathcal{X})=\frac{1}{\sqrt{D}}\sum_{n=1}^{n_x} \left(W_1 x_n\right) \circ \left(W_2 x_n\right)$
\end{algorithmic}
\end{algorithm}

\section{Related Work}
We review the existing studies regarding bilinear pooling in this section.
We first explain the original bilinear pooling, followed by its compact approximation and the meaning of approximation.
Finally, we describe the variants that apply matrix square root normalization to match the scale of the feature elements.

When $d$-dimensional local descriptors $\mathcal{X}=\{x_n\}_{n=1}^{n_x}$ are available, where $x_n\in\mathbb{R}^d$, bilinear pooling \cite{lin2015bilinear} uses second-order statistics
\begin{equation}
 C_x = \sum_{n=1}^{n_x} x_n x_n^t,\label{eq:bilinear}
\end{equation}
as a global image feature.
Because \eqref{eq:bilinear} can be calculated using matrix multiplication, it is easy to calculate the differential; thus, we can learn the whole feature extraction network using backpropagation.
The feature dimension of bilinear pooling is $O(d^2)$, which is large. Therefore, Gao \etal proposed its compact approximation called compact bilinear pooling \cite{gao2016compact} using techniques for approximating polynomial kernels, called tensor sketch \cite{pham2013fast} and random Macraulin \cite{kar2012random}.
This approximation implies the approximation in expectation. In other words, when we have another local descriptor $\mathcal{Y}=\{y_n\}_{n=1}^{n_y}$, where $y_n\in\mathbb{R}^d$ and its bilinear feature $C_y$ and a feature function $\phi$ exists that maps a set of local features to $\mathbb{R}^D$ parameterized by $\theta$, in addition to the distribution $p$ of $\theta$ that satisfies
\begin{equation}
 \langle C_x, C_y \rangle = E_{\theta \sim p}[\langle \phi_\theta(\mathcal{X}),\phi_\theta(\mathcal{Y}) \rangle],\label{eq:approx}
\end{equation}
we can sample $\theta$ from $p$ and use $\phi_\theta(\mathcal{X})$ as an approximation for $C_x$. We can obtain the compact feature by setting $D < d^2$.
We describe the algorithm for compact bilinear pooling using both tensor sketch and random Macraulin in Algorithm \ref{alg:ts}, \ref{alg:rm}, where $\mathrm{fft}$ implies a fast Fourier transform and $\circ$ implies an element-wise product. Both $\phi^{\mathrm{TS}}$ and $\phi^{\mathrm{RM}}$ satisfy the above equality.
Our goal is to obtain $\phi_\theta$ and $p$ that approximate $\langle q(C_x), q(C_y) \rangle$ for any polynomial $q$.

Because a bilinear feature is a second-order statistics, improved bilinear pooling \cite{lin2017improved} applies matrix square root normalization to the bilinear feature $C_x$ and uses $C_x^{1/2}$ as a global feature to alleviate the effect of feature elements with large values and cause the scale of each feature element to be similar. This normalized feature is more compatible with linear classifiers and thus results in higher accuracy. To calculate the backpropagation of a matrix square root, improved bilinear pooling calculates the strict square root using SVD or the approximate square root using Newton--Schulz Iteration \cite{higham2008functions} for forward calculation and solve the Lyapunov equation using SVD for the backward calculation. Liu \etal \cite{li2017second} proposed a direct backward calculation using the differential with respect to singular values and singular vectors. Furthermore, Liu \etal indicated that matrix square root normalization is the robust estimator of a covariance feature as well as the approximation of the Riemannian geometry in the space of positive definite matrices. G2DeNet \cite{wang2017g2denet} used both first- and second-order statistics as a global feature to represent information corresponding to a Gaussian distribution. Because these methods use SVD for the backward calculation, they suffer from computation overhead for GPU calculations and instability when the singular values are similar. Monet \cite{gou2018monet} proposed a compact approximation for improved bilinear feature. Given the G2DeNet feature $M$, Monet first calculated the matrix $Y$ that satisfied $M^{1/2}=Y^t Y$ using SVD. This $Y$ exhibits the property where bilinear features calculated from the row vectors of $Y$ become $M^{1/2}$. Therefore, compact bilinear pooling calculated from row vectors of $Y$ approximates $M^{1/2}$. However, Monet required SVD for forward calculation with respect to $Y$. Li \etal \cite{li2018towards} proposed a method iSQRT-COV that used Newton--Schulz Iteration for both forward and backward calculations. We can perform forward and backward calculations using GPU-friendly matrix multiplications to yield fast computation and improvement in accuracy.

We summarize the comparison of our method and existing methods in \tabref{tab:property}.

\begin{table}[t]
  \caption{Comparison of extended bilinear pooling methods.}
  \label{tab:property}
  \centering
  \begin{tabular}{llll}
    \toprule
    Method & Compact & Matrix Normalization & Avoid SVD \\ \hline\hline
Compact Bilinear Pooling \cite{gao2016compact}& $\checkmark$ & $\times$ & $\checkmark$ \\
Improved Biilnear Pooling \cite{lin2017improved} & $\times$ & $\checkmark$ & $\times$ \\
Monet \cite{gou2018monet} & $\checkmark$ & $\checkmark$ & $\times$ \\
iSQRT-COV \cite{li2018towards} & $\times$ & $\checkmark$ & $\checkmark$ \\
Our work & $\checkmark$ & $\checkmark$ & $\checkmark$ \\
    \bottomrule
  \end{tabular}
\end{table}

\section{Proposed Method}
We describe the proposed method in this section.
Because matrix square root normalization is approximated by the polynomial of the input covariance matrix, in \secref{sec:polynomial}, we propose the approximation for any polynomial of covariance.
In \secref{sec:sqrt}, we combine our proposed method to Newton--Schulz Iteration to obtain the approximation for matrix square root normalization with a small number of iterations.

\subsection{Approximation for polynomial of bilinear feature}\label{sec:polynomial}
Given the local features $\mathcal{X}=\{x_n\}_{n=1}^{n_x}$, $\mathcal{Y}=\{y_n\}_{n=1}^{n_y}$, bilinear feature $C_x, C_y$ was calculated using \eqref{eq:bilinear} and the polynomial $q$; our goal is to calculate the feature function $\phi_{\theta}$ and the distribution $p$ of $\theta$ that satisfies
\begin{equation}
 \langle q(C_x), q(C_y) \rangle = E_{\theta \sim p}[\langle \phi^q_\theta(\mathcal{X}),\phi^q_\theta(\mathcal{Y}) \rangle].\label{eq:polynomialapprox}
\end{equation}
This equation is an extension of \eqref{eq:approx}, where we apply $q$ on the left-hand side.
In our experiment, we used the polynomial of the covariance instead of the polynomial of the bilinear feature to match the setting with iSQRT-COV. We can calculate the approximation by substituting $\frac{1}{\sqrt{n_x}}\left(x_n-\frac{1}{n_x}\sum_{n=1}^{n_x}x_n\right)$ for $x_n$. Furthermore, we can approximate the Gaussian embedding such as G2DeNet by substituting $\frac{1}{\sqrt{n_x}}\begin{pmatrix}x_n^t & 1\end{pmatrix}^t$ for $x_n$.

Our primary result is as follows.
\begin{prop}
We assume that $q$ is written as $q(x) = b + x r(x)$, where $b$ is a bias term and $r$ is a polynomial; 
$\langle x_m x_n^t, y_o y_p^t \rangle$ is approximated as a bilinear form of $\psi_\theta$, written as 
\begin{equation}
 \langle x_m x_n^t, y_o y_p^t \rangle = E_{\theta \sim p}[\langle U \left(\psi^1_\theta(x_m) \otimes \psi^2_\theta(x_n) \right),  U \left(\psi^1_\theta(y_o) \otimes \psi^2_\theta(y_p) \right) \rangle],\label{prop:assumption}
\end{equation}
for $1 \leq m, n \leq n_x$ and $1 \leq o, p \leq n_y$, where $\otimes$ denotes the Kronecker product and $U$ is the deterministic weight matrix.
Furthermore, we denote the matrices that we concatenate, $\psi^i_{\theta}(x_{n})$ and $\psi^i_{\theta}(y_{n})$, as $\Psi^i\left(\mathcal{X}\right) \in \mathbb{R}^{D\times n_x}$ and $\Psi^i\left(\mathcal{Y}\right) \in \mathbb{R}^{D \times n_y}$, respectively.
Then, it follows that 
\begin{align}
& \langle q(C_x), q(C_y) \rangle \nonumber\\ =& E_{\theta \sim p}[\langle c \!+\!\! \sum_{n_1=1}^{n_x} \!\!U\! \left(\!\left(\Psi^1_\theta(\mathcal{X}) r\left(S^x\right)\right)_{:,n_1} \otimes \psi^2_\theta(x_{n_1}) \!\right), c \!+\!\! \sum_{n_1=1}^{n_y} \!\!U\! \left(\!\left(\Psi^1_\theta(y_{n_1})r\left(S^y\right)\right)_{:,n_1} \otimes \psi^2_\theta(y_{n_1}) \!\right)\rangle],\label{polynomial_final}
\end{align}
where $A_{:,n}$ is the $n$-th column vector for matrix $A$; $c$ is a vector calculated using $d$-dimensional element vectors $e_i$ as $c = b \sum_{i=1}^d U\left(\psi^1_\theta(e_i) \otimes \psi^2_\theta(e_i)\right)$ and $S^x_{n_1 n_2} = \langle x_{n_1}, x_{n_2} \rangle$, $S^y_{n_1 n_2} = \langle y_{n_1}, y_{n_2} \rangle$.
\end{prop}
From this proposition, we can calculate the approximate feature as follows:
\begin{equation}
\phi^q_\theta(\mathcal{X})=c + \sum_{n_1=1}^{n_x} U \left(\left(\Psi^1_\theta(\mathcal{X}) r\left(S^x\right)\right)_{:,n_1} \otimes \psi^2_\theta(x_{n_1}) \right).  
\end{equation}
In addition to the feature calculated by compact bilinear pooling, our feature exploits the constant vector $c$ that corresponds to the bias term and the approximate features that correspond to the pair of local features with the weight $r\left(S^x\right)_{n_1 n_2}$.

Before we proceed to the proof, we present some noteworthy points.
First, we can decompose $q(x) = b + x r(x)$ for any polynomial $q$ by setting $b$ as the constant term of $q$ and $r$ as the terms with degree higher than 0 divided by $x$. \eqref{prop:assumption} is the generalization of the tensor sketch and random Macraulin. For example, we derive the random Macraulin from this equation by setting $\psi^i(x)=W_ix$ and $U$ as a matrix corresponding to the mapping from the Kronecker product to the element-wise product divided by $\sqrt{D}$. Additionally, we derive the tensor sketch by setting $\psi^i(x)=\mathrm{fft}\left(\psi^{\mathrm{TS}}(x_n,h_i,s_i)\right)$ and $U$ as the composition of the matrix that corresponds to the fast Fourier transform and the matrix that corresponds to the mapping from the Kronecker product to the element-wise product. The implementation follows the original approximation method. We make this assumption to avoid the calculation of approximate features for all the $n^2$ pair of local features by exploiting bilinearity.
Furthermore, because the summation weight $r(S^x)$ does not depend on $\theta$ and is fixed, the variance and concentration inequality of the proposed feature can be calculated straightforwardly from those of the original approximated feature.

\begin{proof}
First, the monomial of $C_x$ can be represented using $x_n$ and $S^x$ as follows:
\begin{lemm} when we assume $n>0$, it follows that
\begin{equation}
C_x^n = \sum_{n_1=1}^{n_x}\sum_{n_2=1}^{n_x} \left(S^x\right)^{n-1}_{n_1,n_2} x_{n_1} x_{n_2}^t,\label{eq:cxn}
\end{equation}
 \end{lemm}
This lemma can be proved by induction with respect to $n$.
From this lemma, we can prove the case for the monomials.
\begin{lemm}
 When we assume \eqref{prop:assumption} and $n, m > 0$, it follows that 
\begin{align}
 \langle C_x^m, C_y^n \rangle &= E_{\theta \sim p}[\langle \sum_{n_1=1}^{n_x}\sum_{n_2=1}^{n_x} \left(S^x\right)_{n_1 n_2}^{m-1} U \left(\psi^1_\theta(x_{n_1}) \otimes \psi^2_\theta(x_{n_2}) \right), \nonumber\\ & \sum_{n_1=1}^{n_y}\sum_{n_2=1}^{n_y} \left(S^y\right)_{n_1 n_2}^{n-1} U \left(\psi^1_\theta(y_{n_1}) \otimes \psi^2_\theta(y_{n_2}) \right)\rangle],
\end{align}
\end{lemm}
This lemma follows from the fact that 
\begin{equation}
\langle C_x^m, C_y^n \rangle = \sum_{n_1=1}^{n_x}\sum_{n_2=1}^{n_x}\sum_{n_3=1}^{n_y}\sum_{n_4=1}^{n_y} \left(S^x\right)^{n-1}_{n_1,n_2} \left(S^y\right)^{m-1}_{n_3,n_4} \langle x_{n_1} x_{n_2}^t, y_{n_3} y_{n_4}^t\rangle,
\end{equation}
and \eqref{prop:assumption}.
Next, we calculate the case when one term is constant.
\begin{lemm}
 When we assume \eqref{prop:assumption} and $m > 0$, it follows that
\begin{equation}
 \langle C_x^m, I_{d} \rangle = E_{\theta \sim p}[\langle \sum_{n_1=1}^{n_x}\sum_{n_2=1}^{n_x} \left(S^x\right)_{n_1 n_2}^{m-1} U \left(\psi^1_\theta(x_{n_1}) \otimes \psi^2_\theta(x_{n_2}) \right), \sum_{j=1}^d U\left(\psi^1_\theta(e_i)\otimes \psi^2_\theta(e_i)\right)\rangle],
\end{equation}
\end{lemm}
where $I_d$ is the $d$-dimensional identity matrix. This follows from the fact that $I_d$ is a bilinear feature of $e_i$s because $I_d = \sum_{i=1}^d e_i e_i^t$.
Finally, we calculate the case when both terms are constant.
\begin{lemm}
It follows that
\begin{equation}
 \langle I_{d}, I_{d} \rangle = E_{\theta \sim p}[\sum_{j=1}^d U\left(\psi^1_\theta(e_i)\otimes \psi^2_\theta(e_i)\right), \sum_{j=1}^d U\left(\psi^1_\theta(e_i)\otimes \psi^2_\theta(e_i)\right)\rangle],
\end{equation}
\end{lemm}
Because the general polynomial $q$ can be calculated as the sum of monomials, we can summarize the results of the above lemmas to obtain \eqref{polynomial_final}.
\end{proof}

From the above proposition and 
\begin{equation}
\sum_{n_1=1}^{n_x}\sum_{n_2=1}^{n_x} r\left(S^x\right)_{n_1 n_2} U \left(\psi^1_\theta(x_{n_1}) \otimes \psi^2_\theta(x_{n_2}) \right)=\sum_{n_1=1}^{n_x} U \left(\left(\Psi^1_\theta(\mathcal{X}) r\left(S^x\right)\right)_{:,n_1} \otimes \psi^2_\theta(x_{n_1}) \right),
\end{equation}
because of the bilinearity of the Kronecker product, we obtained the approximated feature for the general polynomial.

\subsection{Application to matrix square root normalization}\label{sec:sqrt}
In this section, we apply the approximation method obtained in \secref{sec:polynomial} for matrix square root normalization.
Our strategy is to first calculate $b, r(S^x)$ that corresponds to the 1/2 power of $S^x$ and then substitute these matrices in \eqref{polynomial_final}.
Interestingly, we can obtain $b, r(S^x)$ iteratively without calculating the explicit form of the polynomial that corresponds to 1/2 power.

We first review Newton--Schulz iteration.
Given $d\times d$ matrix $A$ that satisfies $\|A-I_d\|<1$,
Newton--Schulz iteration sets $Y_0=A, Z_0=I_d$ and then calculates $Y_k, Z_k$ iteratively as follows:
\begin{align}
 Y_{k+1} &= \frac{1}{2} Y_k \left(3 I_d - Z_k Y_k\right) \\
 Z_{k+1} &= \frac{1}{2} \left(3 I_d - Z_k Y_k\right) Z_k.
\end{align}
It is known that $Y_k$ converges to $A^{1/2}$.
Furthermore, as shown in this equation, $Y_{k}, Z_{k}$ are calculated using only matrix multiplication and summation from $A$ and $I_d$. This implies that $Y_k, Z_k$ are obtained as the polynomials of $A$.
Moreover, when we view $Y_k$ as the polynomial of $A$, $Y_0 = A$ and $Y_{k+1}$ is the multiple of $Y_k$. Thus, the constant term of $Y_k$ is 0. Thus, $b=0$ holds.
Furthermore, we can prove by induction that $Y_k = A Z_k$. This implies that $r(A)$ is obtained as $Z_k$,
and is contrary to the original iSQRT-COV that uses $Y_k$ as the feature.

For a general $S^x$, as in the case of iSQRT-COV, we normalize $S^x$ as $\frac{1}{\mathrm{trace}\left(S^x\right)}S^x$ such that the norm of difference to the identity matrix is smaller than 1. Then, we calculate $r(S^x)$ as $\frac{1}{\sqrt{\mathrm{trace}\left(S^x\right)}}Z_k$.
The calculation of trace requires only matrix multiplication and summation.
We summarize our algorithm in \algref{alg:proposed}.
We call our method improved polynomial compact covariance pooling (iPCCP).

\begin{algorithm}[t]
\caption{Calculating the approximation for matrix square root of bilinear feature}         
\label{alg:proposed}                          
\begin{algorithmic}                  
\REQUIRE $\mathcal{X}=\{x_n\}_{n=1}^{n_x}$, where $x_n\in\mathbb{R}^d$, feature dimension $D$, iteration number $k$
\ENSURE $\phi(\mathcal{X}) \in \mathbb{R}^D$
\STATE Calculate $S^x$ where $S^x_{n_1 n_2} = \langle x_{n_1}, x_{n_2} \rangle$
\STATE  $A = \frac{1}{\mathrm{trace}{S^x}}S^x$
\STATE $Y_0 = A, Z_0 = I_d$
\FOR{$i=0$ to $k-1$}
\STATE $Y_{k+1} = \frac{1}{2} Y_k \left(3 I_d - Z_k Y_k\right)$
\STATE $Z_{k+1} = \frac{1}{2} \left(3 I_d - Z_k Y_k\right) Z_k$
\ENDFOR
\STATE $S=\frac{1}{\sqrt{\mathrm{trace}{S^x}}} Z_k$
\STATE $\phi(\mathcal{X})= \sum_{n_1=1}^{n_x} U \left(\left(\Psi^1_\theta(\mathcal{X})S\right)_{:,n_1} \otimes \psi^2_\theta(x_{n_1}) \right)$
\end{algorithmic}
\end{algorithm}

\subsection{Computational Complexity}\label{sec:complexity}
In this section, we calculate the complexity of the proposed pooling method.
We assume the local feature dimension as $d$, the number of local feature as $n$, and the global feature dimension as $D$.
When we use the tensor sketch and random Macraulin approximations, the computation of $\Psi^i_\theta\left(\mathcal{X}\right)$ requires $O\left(nd + nD\log D\right)$ and $O(ndD)$, respectively. Furthermore, the computation of $S$ requires $O(n^2d + n^3)$, and $\Psi^1_\theta(\mathcal{X})S$ requires $O(n^2D)$.
The total complexity is $O\left(nD\log D + n^2d + n^3 + n^2D\right)$ for the tensor sketch and $O(ndD + n^2d + n^3 + n^2D)$ for the random Macraulin. In ordinary cases, we can assume $d \simeq n \ll D < d^2 \simeq n^2$. Therefore, the computational complexities for both the pooling methods are equivalent to $O(n^2D)$ for the tensor sketch and $O(ndD + n^2D)$ for the random Macraulin. Owing to the calculation of $S$, iPCCP with the tensor sketch approximation is slightly slower than the compact bilinear pooling\cite{gao2016compact} 
utilizing the same approximation $O(nd + nD\log D)$. Meanwhile, when the random Macraulin approximation is employed, the computational complexity of iPCCP is asymptotically equal to that of the compact bilinear pooling $O(ndD)$.

\begin{figure}[t]
\centering
\subfloat[][CUB]{\includegraphics[width=0.32\hsize]{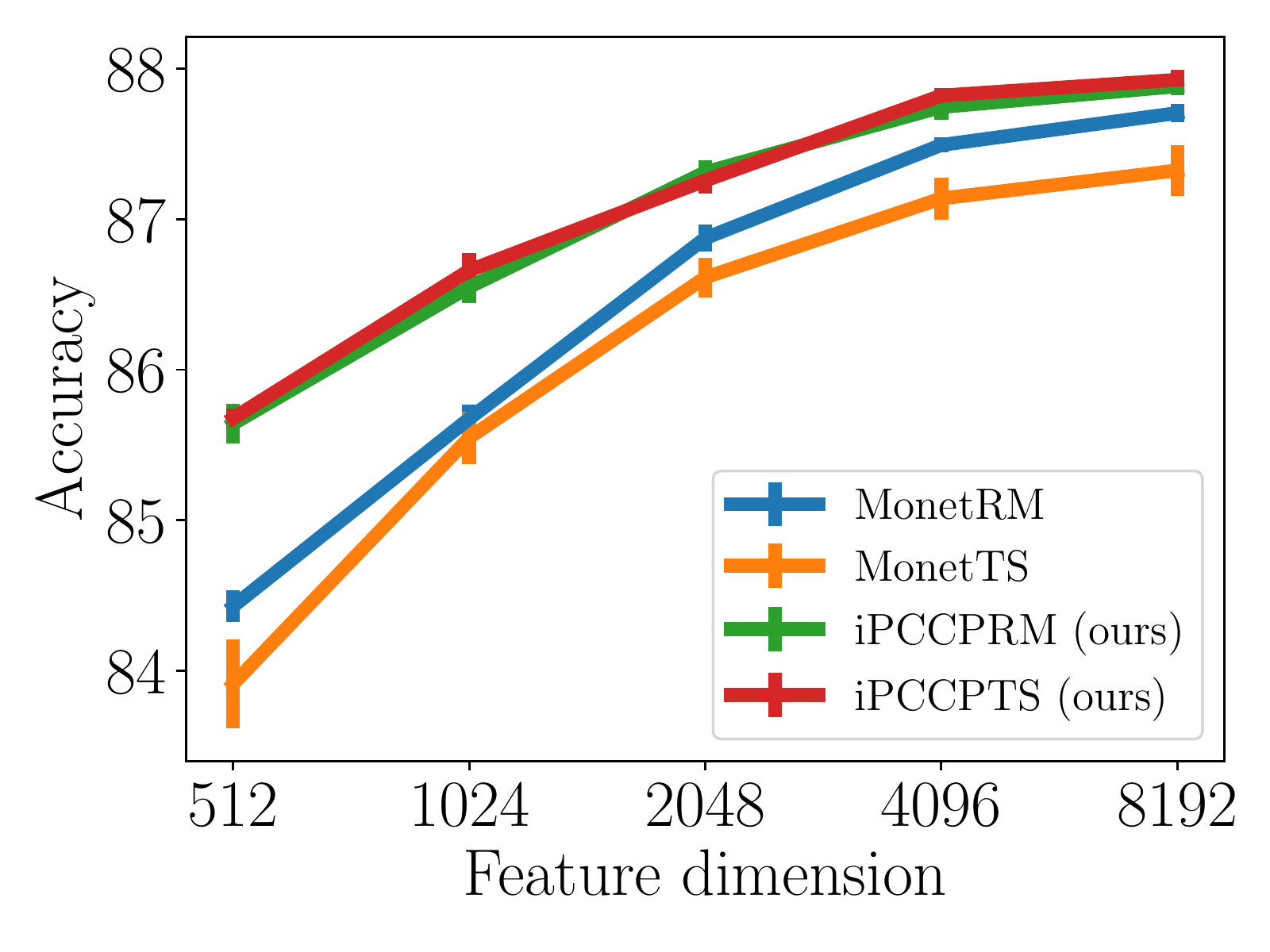}}
\subfloat[][Aircraft]{\includegraphics[width=0.32\hsize]{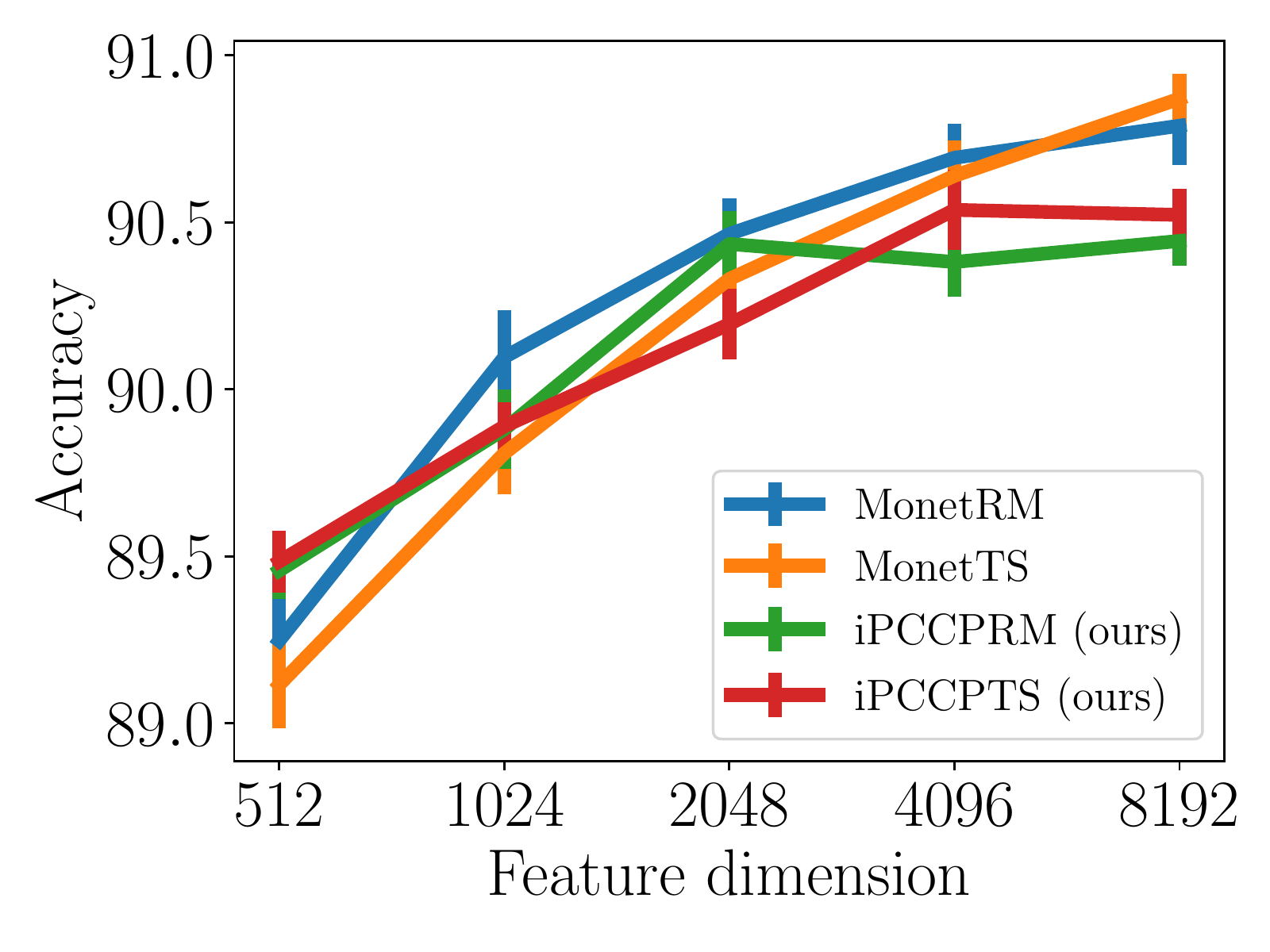}}
\subfloat[][Cars]{\includegraphics[width=0.32\hsize]{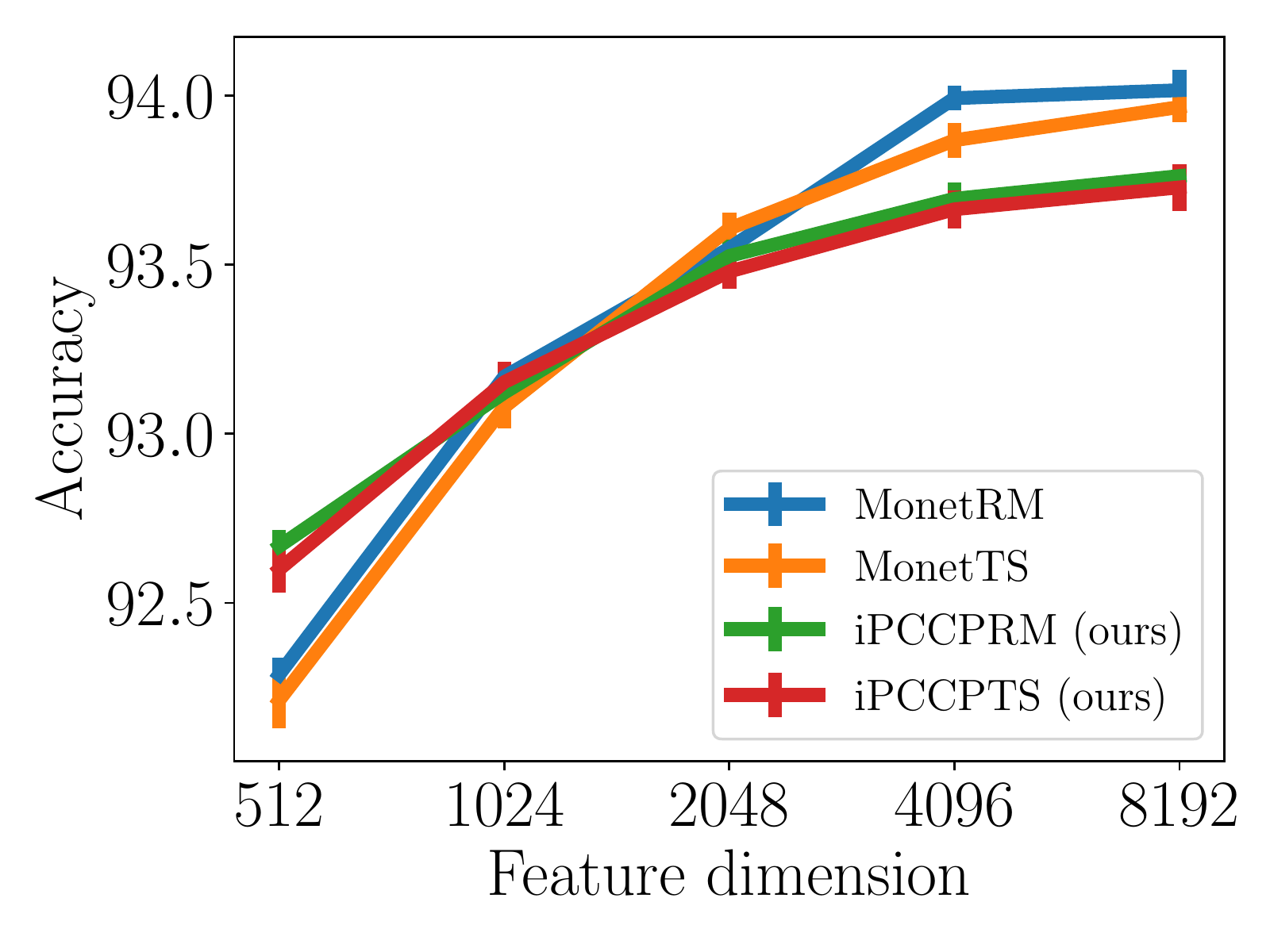}}
\vspace{-3mm}
  \caption{Accuracy using Resnet50 architecture.}\label{expr:res50}
\vspace{-5mm}
\end{figure}

\begin{figure}[t]
\centering
\subfloat[][CUB]{\includegraphics[width=0.32\hsize]{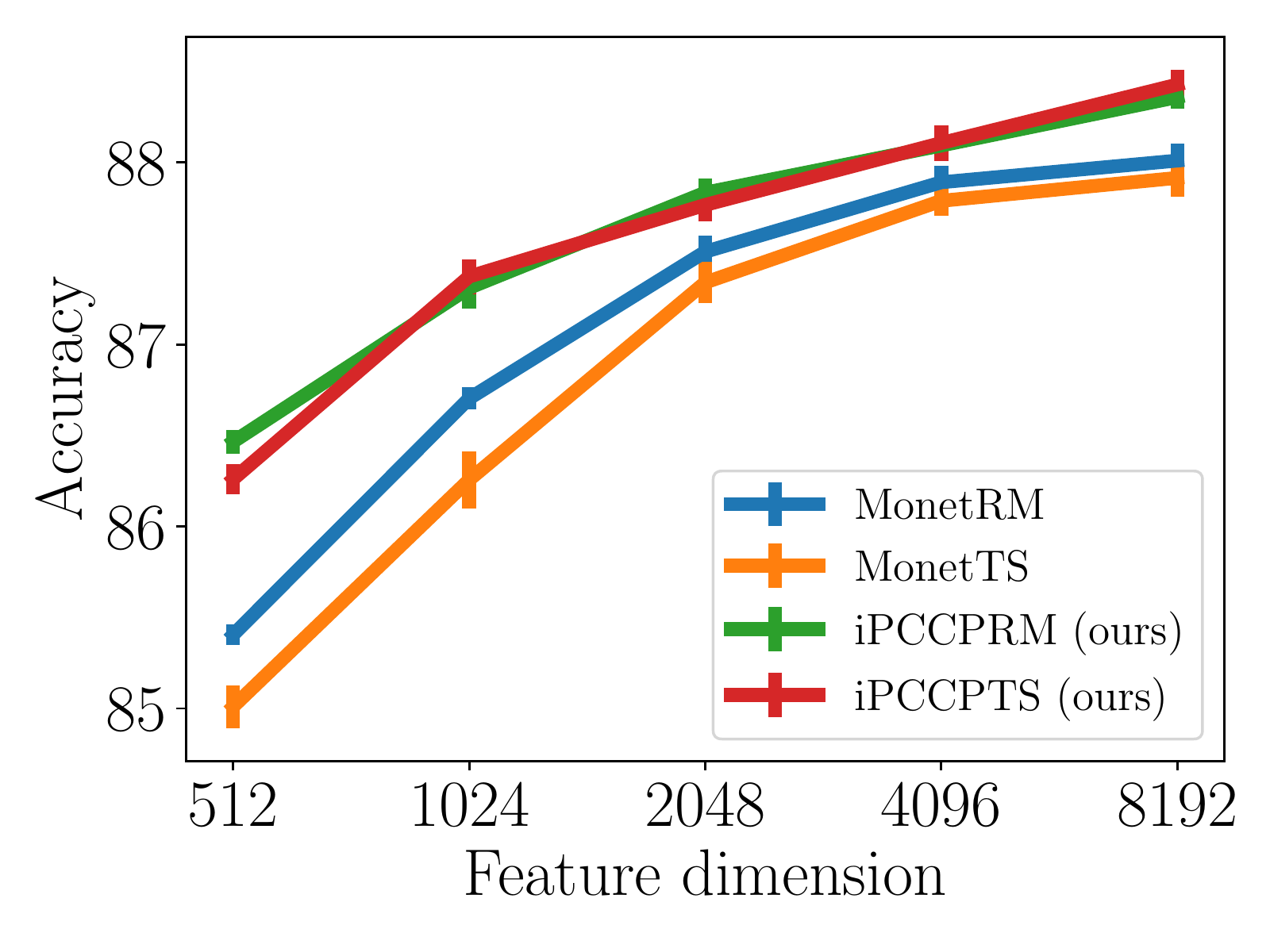}}
\subfloat[][Aircraft]{\includegraphics[width=0.32\hsize]{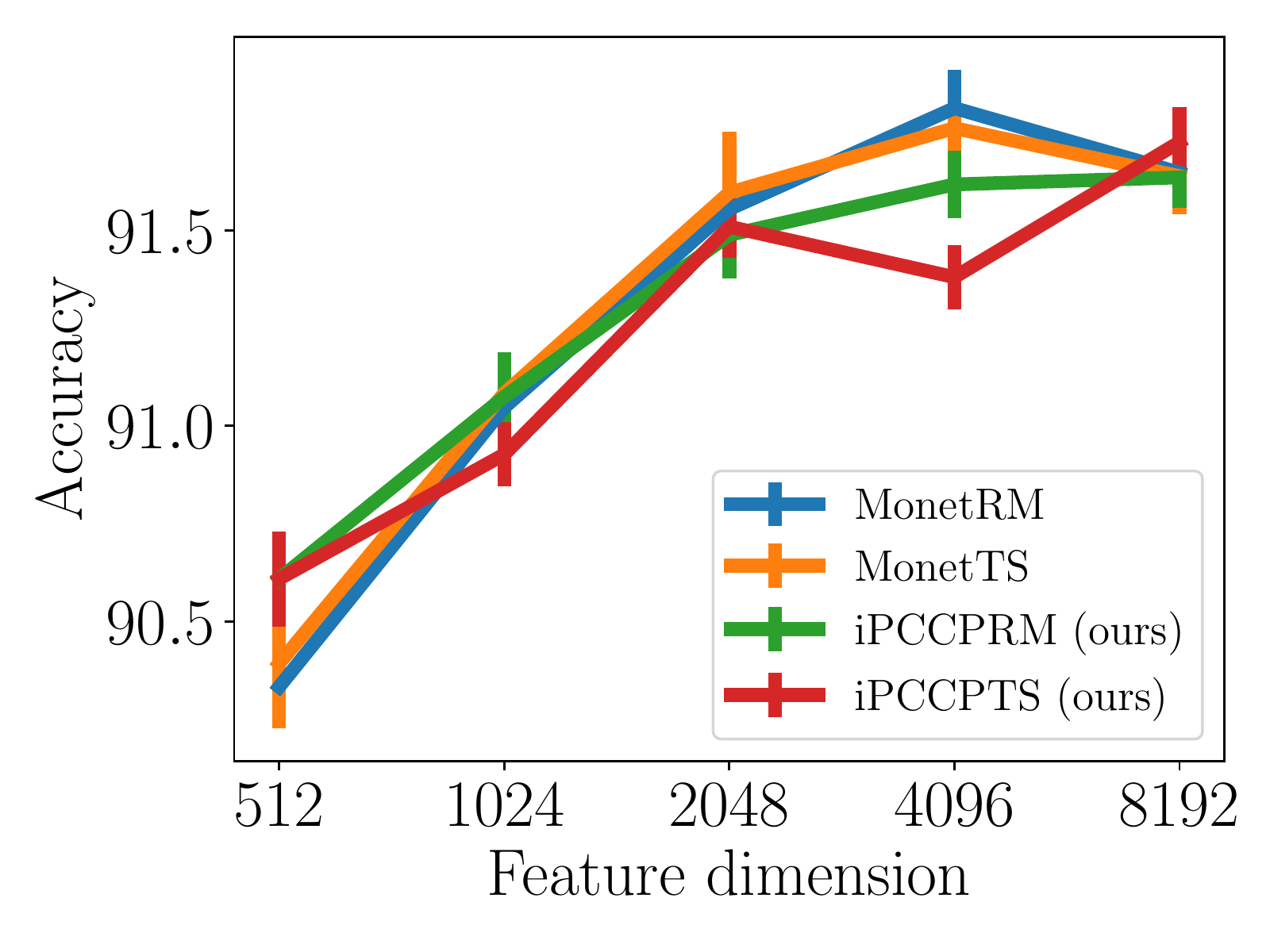}}
\subfloat[][Cars]{\includegraphics[width=0.32\hsize]{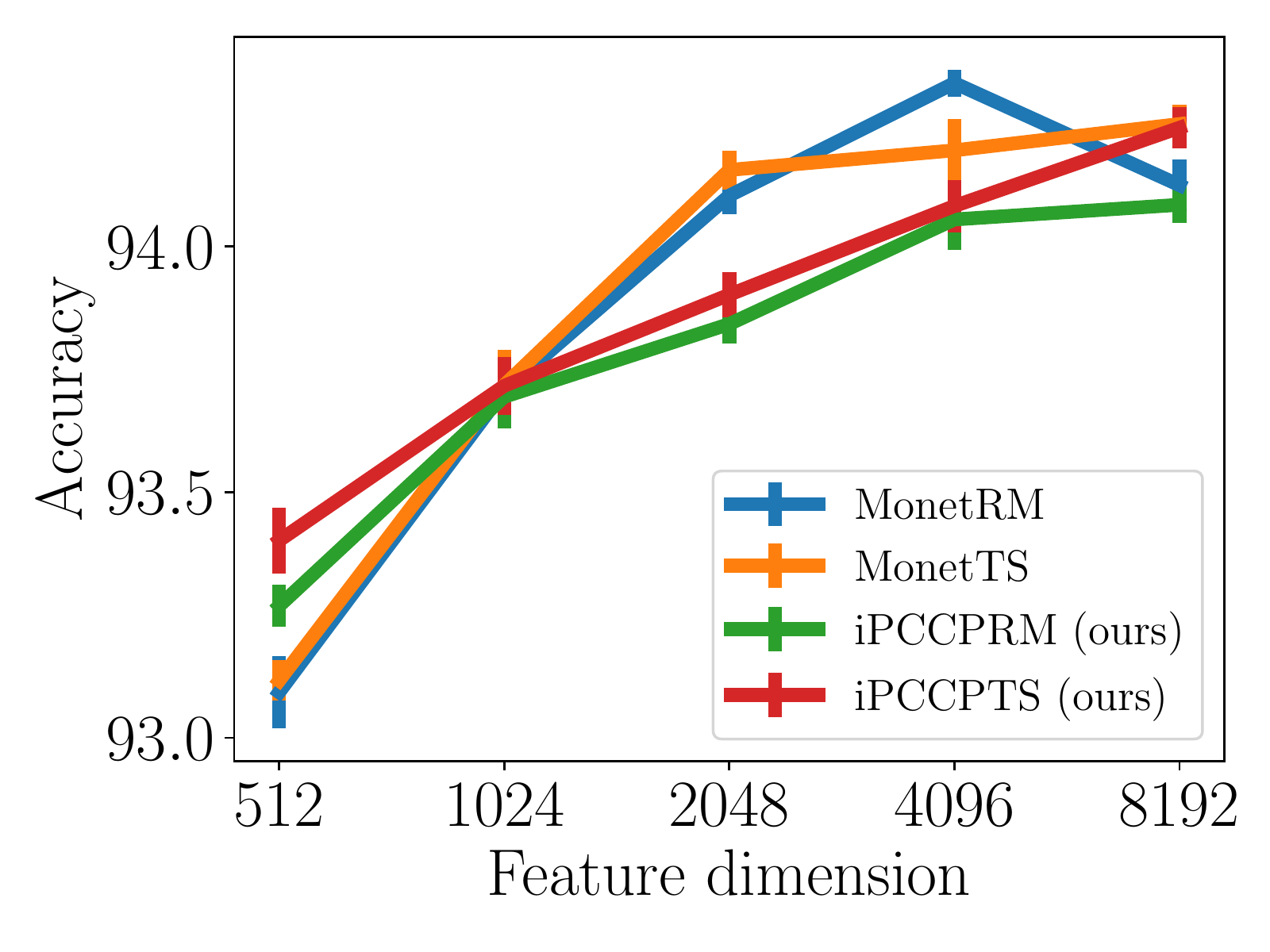}}
\vspace{-3mm}
  \caption{Accuracy using Resnet101 architecture.}\label{expr:res101}
\vspace{-5mm}
\end{figure}

\section{Experiment}
In this section, we evaluate the classification accuracy of the proposed method on standard fine-grained image recognition datasets.

We evaluate the methods on the Caltech-UCSD Birds (CUB) \cite{welinder2010caltech}, FGVC-Aircraft Benchmark (Aircraft) \cite{maji13fine-grained}, and Stanford Cars (Cars) datasets \cite{krause20133d}.
The CUB dataset is a dataset consisting of 200 categories bird images with 5,994 training images and 5,794 test images.
The Aircraft dataset is a fine-grained airplane image dataset of 100 categories with 6,667 training images and 3,333 test images.
The Cars dataset consists of 196 categories of car images with 8,144 training images and 8,041 test images.

We used pretrained Resnet50 and 101 \cite{he2016deep} architectures as local feature extractors.
To obtain pretrained models that are appropriate for covariance pooling, we pretrained the model that replaced global average pooling with iSQRT-COV \cite{li2018towards}. The pretrained setting of iSQRT-COV Resnet was based on \cite{li2018towards}. We used the ILSVRC2012 dataset \cite{imagenet_cvpr09} with random clipping and random horizontal flipping augmentations. We set the batch size to 160 and weight decay rate to 1e-4, and used a momentum grad optimizer with an initial learning rate 0.1 and momentum 0.9 for 65 epochs. We multiplied the learning rate by 0.1 at 30, 45 and 60 epochs.

Based on the pretrained model, we changed the pooling method to iSQRT-COV, Monet, and iPCCP, and compared the accuracy of the fine-tuned models.
To match the local features, we used Monet on the covariance. This corresponds to Monet-2 in \cite{gou2018monet}.
We varied the approximation method to random Macraulin and tensor sketch, and approximated the dimensions as 512, 1,024, 2,048, 4,096, and 8,192 for Monet and iPCCP.

Following the settings of the existing studies, we resized each input image to 448 x 448 for fine-tuning and evaluation. However, we resized the images to 512 x 512 and cropped the 448 x 448 center image for the Aircraft dataset.
Therefore, we obtain a convolutional feature of 28 x 28 x 2,048 dimension as the local feature. We then applied a 1 x 1 convolution layer to reduce the dimension to 28 x 28 x 256 and applied the pooling methods.
We applied random horizontal flipping augmentation for fine-tuning, and used the average score for the test image and flipped the test image for evaluation.
We set the batchsize as 10 and weight decay rate as 1e-3, and learned the model with momentum grad with learning rates of 1.2e-3 for the feature extractor and 6e-3 for the classifier. We learned the model for 100 epochs.

For each setting, we evaluated the models ten times and evaluated the mean and standard error of the test accuracy.

\begin{table}[t]
\centering
\caption{Accuracy comparison with iSQRT-COV. The score with ‘‘(reported)'' denotes the score reported in the original paper \cite{li2018towards}.}
\label{table:expr}
\small
{\tabcolsep=1.5pt
 \begin{tabular}[t]{|c|c|c|c|c|c|c|c|}\hline
Method & Dim & \multicolumn{3}{|c|}{Resnet50} & \multicolumn{3}{|c|}{Resnet101} \\\hline
  & &CUB &Aircraft &Cars &CUB &Aircraft &Cars \\\hline
iSQRT-COV (reported) \cite{li2018towards} &32k &88.1 &90.0 &92.8 &88.7 &91.4 & 93.3\\
iSQRT-COV \cite{li2018towards} &32k & 87.96$\pm$0.03&90.69$\pm$0.08&93.62$\pm$0.03&88.08$\pm$0.07&92.05$\pm$0.07&93.96$\pm$0.05 \\
Monet-TS \cite{gou2018monet} &8k & 87.32$\pm$0.17&90.87$\pm$0.08&93.97$\pm$0.04&87.92$\pm$0.10&91.64$\pm$0.10&94.25$\pm$0.04 \\
Monet-RM \cite{gou2018monet} &8k &87.71$\pm$0.06&90.79$\pm$0.12&94.02$\pm$0.06&88.01$\pm$0.09&91.65$\pm$0.09&94.12$\pm$0.05 \\
iPCCP-TS (ours) &8k &87.93$\pm$0.07&90.52$\pm$0.08&93.73$\pm$0.07&88.43$\pm$0.08&91.73$\pm$0.09&94.24$\pm$0.04 \\
iPCCP-RM (ours) &8k &87.88$\pm$0.06&90.44$\pm$0.07&93.76$\pm$0.03&88.36$\pm$0.06&91.64$\pm$0.08&94.08$\pm$0.04 \\ \hline
 \end{tabular}
}
\end{table}

\begin{table}[t]
\centering
\vspace{-5mm}
\caption{Comparison time (s) to calculate training/evaluation computation for one minibatch.}
\label{table:time}
\small
 \begin{tabular}[t]{|c|c|c|c|}\hline
Network & iSQRT-COV & Monet-RM & iPCCP-RM (ours) \\\hline
Resnet50 & 0.16 / 0.05 & 0.59 / 0.52 & 0.24 / 0.08 \\
Resnet101 & 0.23 / 0.08 & 0.71 / 0.53 & 0.32 / 0.10 \\ \hline
 \end{tabular}
\end{table}

Figures \ref{expr:res50} and \ref{expr:res101} show the results. 
It is noteworthy that the y-axis scale of the CUB is approximately twice as large as that of the other datasets.
The proposed method demonstrates comparable performance to Monet. This is because these two methods approximate the same feature.
However, our iPCCP demonstrates better performance when the feature dimension is 512 and on the CUB dataset, which dataset demonstrates the lowest accuracy and is thus the most difficult to recognize among the datasets used.
We can avoid SVD and approximate the covariance stably using our method; this contributes to learning the discriminative feature even for difficult settings. Furthermore, we  compared the score for the 8,192-dimensional feature with iSQRT-COV. \tabref{table:expr} shows that our methods obtain scores similar to that using iSQRT-COV in all the settings with approximately 1/4 feature dimension.

Furthermore, we compared the computation time. 
Because the tensor sketch approximation is more complex and the computation time varies significantly according to the implementation, we evaluated the time for the random Macraulin approximation.
We compared iSQRT-COV with Monet and the proposed IPCCP for a 8,192-dimensional feature.
We implemented each method using pytorch on NVIDIA Tesla V100 and compared the time to calculate the training/evaluation for one minibatch.

\tabref{table:time} shows that iPCCP-RM requires a longer computation time than iSQRT-COV arising from the additional computation described in \secref{sec:complexity}. This additional time is almost the same as the overhead to change the architecture from Resnet50 to Resnet101 and, thus, is not too large to incur difficulty in training and evaluation. Monet required more than twice as long a time compared with iPCCP for training and five times for evaluation.
That Monet required a significant amount of time for both training and evaluation could be attributed to the SVD required for forward computation.

The above results demonstrate that the proposed iPCCP is an effective approximation method for an improved covariance feature.

\section{Conclusion}
We herein proposed a novel direct approximation for covariance features with matrix square root normalization.
We applied the fact that the recently proposed Newton--Schulz Iteration-based method approximates the matrix square root as the polynomial of an input covariance matrix, and constructed the approximation for the matrix square root from the approximation for the general polynomial of a covariance matrix.
We evaluated the proposed approximation method on fine-grained image recognition datasets and demonstrated a similar accuracy compared with the original feature with fewer dimensions and less time than the existing approximation methods.

\subsubsection*{Acknowledgements}
This work was partially supported by JST CREST Grant Number JPMJCR1403, Japan, and partially supported by JSPS KAKENHI Grant Number JP19176033.

 {
 \bibliographystyle{ieee}
 \bibliography{neurips_2019_arxiv}
 }

\end{document}